\documentclass{llncs}
\usepackage{tcolorbox}
\usepackage{footmisc}
\usepackage[style=1,leftmargin=0pt,rightmargin=0pt]{mdframed}
\usepackage{multicol}
\usepackage{listings}
\usepackage{amsmath,amssymb}
\usepackage{tikz}
\usepackage{tikz-cd}
\tikzcdset{scale cd/.style={every label/.append style={scale=#1},cells={nodes={scale=#1}}}}

\begin{document}

\title{On the use of neurosymbolic AI for \\ defending against cyber attacks}

\author{
Gudmund Grov\inst{1,2}
\and
Jonas Halvorsen\inst{1}
\and
Magnus Wiik Eckhoff\inst{1,2}
\and
Bjørn Jervell Hansen\inst{1}
\and
Martin Eian\inst{3}
\and
Vasileios Mavroeidis \inst{2}
}

\institute{
Norwegian Defence Research Establishment (FFI), Kjeller, Norway\\
\email{\{Gudmund.Grov,Jonas.Halvorsen,Bjorn-Jervell.Hansen,Magnus-Wiik.Eckhoff\}@ffi.no}\\
\and
University of Oslo, Norway, \email{vasileim@ifi.uio.no}
\and
mnemonic, Norway, \email{meian@mnemonic.no}\\ 
}

\newcommand{\kommentar}[1]{\textcolor{red}{#1}}

\newtheorem{challenge}{Challenge}
\tcolorboxenvironment{challenge}{
  colback=red!5!white,
  boxrule=0pt,
  boxsep=1pt,
  left=2pt,right=2pt,top=2pt,bottom=2pt,
  oversize=2pt,
  before skip=\topsep,
  after skip=\topsep,
}

\newtheorem{usecase}{Use case}
\tcolorboxenvironment{usecase}{
  colback=blue!5!white,
  boxrule=0pt,
  boxsep=1pt,
  left=2pt,right=2pt,top=2pt,bottom=2pt,
  oversize=2pt,
  before skip=\topsep,
  after skip=\topsep,
}

\maketitle

\small

\keywords{AI, neurosymbolic AI, cyber security, incident detection and response}

\begin{abstract}
It is generally accepted that all cyber attacks cannot be prevented, 
creating a need for the ability to detect and respond to cyber attacks.
Both connectionist and symbolic AI are currently being used 
to support such detection and response.
In this paper, we make the case for combining them using neurosymbolic AI. 
We identify a set of challenges when using AI today and propose a set of neurosymbolic use cases we believe are both interesting research directions for the neurosymbolic AI community and can have an impact on the cyber security field.
We demonstrate feasibility through two proof-of-concept experiments.
\end{abstract}

\section{Introduction}

Protecting assets in the cyber domain requires a combination of \emph{preventive measures}, such as access control and firewalls, and the ability to \emph{defend} against cyber operations when the preventive measures were not sufficient.\footnote{It is generally accepted that preventive measures are not sufficient in cyberspace. An analogy is that we still need smoke detectors and a fire brigade, even if we take all possible preventive measures to reduce the risk of a fire.}

Our focus in this paper is on defending against offensive cyber operations, and before going into details, some concepts and terminology need to be in place:

\makeatletter

\begin{mdframed}
[leftmargin=0pt,rightmargin=0pt,backgroundcolor=black!5!white,linecolor=black!5!white,fontcolor=black]
\textbf{Terminology \& concepts}\;\; 
The focus of this paper is to defend \emph{assets} against \emph{threats} in cyberspace. An \emph{asset} can be anything from information or physical infrastructure to the internal processes of an enterprise. 
\emph{Threats} manifest themselves in the form of \emph{cyber operations} (or \emph{cyber attacks}) conducted by an \emph{adversary} (or \emph{threat actor}).
In our context of defending, the term \emph{incident} is used for a potential attack that is deemed to have an impact on assets, and the process of defending comes under the area of \emph{incident management} \cite{cichonski2012computer}.
This is typically conducted in a \emph{security operations centre} (SOC), which consists of 
people, processes and tools. One of the objectives of a SOC is to detect and respond to threats and attacks, where \emph{security analysts} play a key role.
Knowledge/intelligence of threats and threat actors in the cyber domain is called \emph{cyber threat intelligence} (CTI). Networks and systems to be protected are monitored and 
\emph{events} -- e.g., network traffic, file changes, or processes executing on a host -- are forwarded
and typically stored in a \emph{security information and event management} (SIEM) system, where events can be searched and queried. We will use the term \emph{obseverations} for these events resulting from monitoring. Suspicious activity that is observed may raise 
\emph{alerts}, which may indicate an incident that has to be analysed and responded to in the SOC.
\emph{Neurosymbolic AI} \cite{garcez2023neurosymbolic}, which aims to combine connectionist and symbolic AI, will be abbreviated \emph{NeSy}.
\end{mdframed}

Why is a SOC relevant for NeSy? A SOC essentially conducts abductive reasoning by observing traces and identifying and analysing their cause in order to respond. This involves
sifting through masses of events for suspicious behaviour, an area in which there has been extensive research for several decades using statistics and machine learning.
Identifying the cause of observed suspicious behaviour requires situational awareness, achieved by combining and reasoning about different types of knowledge. There are various ways knowledge is represented, such as structured events and alerts, unstructured reports, and semantic knowledge \cite{sikos2023cybersecurity,liu2022review}.
In a SOC, the ability to \emph{learn} models to detect suspicious activities, and the ability to \emph{reason} about identified activity from such models to understand their cause and respond, is thus required.
These abilities are at the core of NeSy and our hypothesis is that 
`\textit{a SOC provides an ideal environment to study NeSy with great potential for both  
 scientific and financial impact'.}
Some early work has explored NeSy in the cyber security domain 
\cite{jalaipan23,piplai2023knowledgeIEEE,himmelhuber2022detection,onchis2022advantages,dingaccelerating24}
and our goal with this paper is to showcase the possibilities and encourage the NeSy community to conduct research in the SOC field. The contributions of the paper are threefold: (1) we outline how AI is used today in a SOC and identify and structure a set of challenges practitioners using AI are faced with; (2) we create a set of promising use cases for NeSy in a SOC and review current NeSy approaches in light of them; (3)  we conduct two proof-of-concept NeSy experiments to showcase feasibility. The focus here is to demonstrate NeSy possibilities and not on realistic conditions or performance.



\paragraph{Methodology:} The identified challenges are derived from a combination of existing published studies of SOCs, the experience and expertise of the authors and discussions with SOC practitioners. The use cases are a result of reviewing NeSy literature in the context of the identified challenges, and the experiments follows from the use cases.

\paragraph{Paper structure:}
In \S \ref{sec:problem}, we describe the typical use of AI in a SOC and the identified challenges. In \S \ref{sec:why}, we make the case for NeSy and outline the NeSy use cases. In 
\S \ref{sec:experiments}, we describe the proof-of-concept experiments, before we conclude in \S \ref{sec:concl}.

\section{Challenges faced when using AI in a SOC}\label{sec:problem}



MAPE-K (Monitor-Analyse-Plan-Execute over a shared Knowledge) \cite{kephart2003vision} is a common reference model to structure the different phases when managing an incident.\footnote{Other common reference models are the OODA (Observe-Orient-Decide-Act) loop \cite{boyd1996essence} and IACD (Integrated Adaptive Cyber Defense) \cite{done2016towards}.} For each phase of MAPE-K, we below discuss the common use of AI, including underlying representations, and identify key challenges security practitioners face when using AI.\footnote{There has
recently been a vast number of proposals for using \emph{large language models} (LLMs) across MAPE-K.
Here, we include what we currently consider the most promising uses of LLMs, and refer to \cite{motlagh2024large,Bodungen24chatgpt} for a more complete discussion.}

\paragraph{Monitor}

In the \emph{monitor} phase, systems and networks are monitored and the telemetry is represented as sequences of \emph{events}.
An \emph{event} could, for instance, be a network packet, a file update, a user that logs on to a service, or a process being executed. Events are typically structured as key-value pairs. For a large enterprise, there may be tens of thousands of events generated per second. In this phase, a key objective is to detect suspicious behaviours from the events
and generate \emph{alerts}, which are analysed and handled in the later phases of MAKE-K.




This is a topic where machine learning (ML) has been extensively studied by training ML models on the vast amount of captured event data  (see e.g. \cite{ahmad2021network}). A challenge with such data is the lack of \emph{ground truth}, in the sense that for the vast majority of events we do not know if they are benign or malicious. As most events will be benign (albeit we do not know which ones), one can exploit this assumption and use unsupervised methods to train anomaly detectors. This is a common approach. For at least research purposes,   
synthetic datasets from simulated attacks are also commonly used \cite{kilincer2021machine}.
However, synthetic datasets suffer from several issues \cite{kenyon2020public,apruzzese2023sok}
and promising results in research papers using synthetic data tend not to be recreated in real-world settings -- whilst anomaly detectors often create a high number of false alerts\footnote{As most events are benign, the base rate fallacy is important in this domain.} \cite{bushra_a_alahmadi_99_2022,sommer2010outside}. Our first challenge, which follows from the \emph{European Union Agency for Cybersecurity} (ENISA) \cite{enisaAI23}, 
addresses this performance issue for ML models for real-world conditions:
\begin{challenge}
\label{challenge:performance}
Achieve optimal accuracy of ML models under real-world conditions.
\end{challenge}
As both normal software and malware are continuously updated, the notion of \emph{concept drift} is prevalent, and ML models must thus be retrained regularly. Moreover, in addition to the large amount of data, requiring scalability,  real-world conditions will have a large amount of noise  (i.e.\ aleatoric uncertainty) in the data, which is not well reflected in synthetic data.

For previous incidents that have been handled, we know the ground truth of the associated alerts and events. Compared with the full set of events, this dataset will, however, be tiny. Still, it is important as it is labelled and contains data we know are relevant -- either in terms of actual attacks experienced or false alerts that should be filtered out. One important challenge, also identified by ENISA \cite{enisaAI23}, is the ability to exploit such labelled  ``incident datasets'' and train ML models based on them:
\begin{challenge}
\label{challenge:smalldata}
Learning with small (labelled) datasets (from cyber incidents). 
\end{challenge}


New knowledge, e.g. about certain threats, attacks, malware or vulnerability exploits, is frequently published (in e.g. threat reports). The traditional, and still most common, method of threat detection is so-called \emph{signature-based} detection, where such knowledge is (often manually) encoded as specific patterns (called signatures). Detection is achieved by matching events with these signatures, and generating alerts when they match.
While signature-based methods have their limitations, such knowledge could improve the performance of ML-based detection models trained on event data, requiring the ability to 
extract  relevant knowledge and including it in the ML models:
\begin{challenge}
\label{challenge:extractk}
Extract knowledge (including about threats, malware and vulnerabilities) and enrich ML-based detection models with it.
\end{challenge}
In addition to reports, there are many knowledge bases and formal ontologies that can be used to enrich ML models with such knowledge.
There are also attempts to extend the coverage domain for such ontologies:
one example is the \emph{unified cybersecurity ontology} (UCO) \cite{syed2016uco} and another is the 
SEPSES knowledge graph \cite{kiesling2019sepses}. 
To represent cyber threat intelligence (CTI), a commonly used schema is \emph{Structured Threat Information Expression} (STIX) \cite{STIX}, with the associated \emph{Threat Actor Context} (TAC) ontology \cite{Vasileios21}.
A widely used knowledge base for threat actors and attacks is the MITRE ATT\&CK  \cite{ATTACK}.
ML, and in particular \emph{natural language processing} (NLP), is being explored for extracting CTI into symbolic forms (e.g. STIX) \cite{marchiori2023stixnet} or to map with MITRE ATT\&CK \cite{li2022attackg}.  \emph{Large language models} (LLMs) are also explored for this topic \cite{liu2023constructing,haque2023adversarial}, with limitations identified \cite{wursch2023llms}. 
However, we are not familiar with approaches to combine and integrate such knowledge with ML models for detection trained on events. 

A different approach to identify malicious behaviour is \emph{cyber threat hunting}. This is a \emph{hypothesis-driven} approach where hypotheses are iteratively formulated (typically using CTI) and validated using event logs, as well as other knowledge \cite{shu2018threat}. Automating this process is our final challenge for the monitor phase:
\begin{challenge}
\label{challenge:hunting}
Automated generation of hypotheses from CTI and validation of hypotheses using observations for threat hunting.
\end{challenge}
\noindent There are AI-based approaches to facilitate threat hunting \cite{nour2023survey}. Most have focused on supporting hypothesis generation by extracting relevant CTI, and using both ML/NLP \cite{gao2021enabling} and symbolic AI  \cite{qamar2017data}. Symbolic AI has been used to support validation \cite{Chetwin24}.







\paragraph{Analyse}

The goal of the analysis phase is to understand the nature of the observed alerts, determine possible business impact and create sufficient situational awareness to support the subsequent pland and execute phases. 

Both malware and benign software continuously evolve. This makes it difficult to 
separate malicious from benign behaviour \cite{enisaAI23}, despite continuous detection 
engineering efforts improving the capabilities. For example, an update to benign software may cause a match with an existing malware signature, and may also show up as an anomaly in the network traffic. As a result, most of the alerts are either false or not sufficiently important for further investigation \cite{bushra_a_alahmadi_99_2022}.
The analysis phase is, therefore, labour-intensive, where security analysts must plough through and analyse a large number of alerts -- most of them false -- to decide their nature and importance. This could lead to so-called \emph{alert fatigue} among security analysts:
\begin{challenge}
\label{challenge:volume}
The volume of alerts leads to alert flooding and alert fatigue in SOCs. 
\end{challenge}
Understanding the nature of alerts is important, and studies have shown that 
a lack of understanding of the underlying scores, or reasoning, behind the alerts 
have led to misuse and mistrust of ML systems \cite{oesch20}. 
Both studies \cite{bushra_a_alahmadi_99_2022} and guidance from ENISA \cite{enisaAI23} have highlighted the need for alerts to be reliable, explainable and possible to analyse. 
The use of explainable AI to support this has shown some promise \cite{eriksson22}, and both knowledge graphs 
\cite{bushra_a_alahmadi_99_2022} and LLMs\footnote{E.g. Microsoft security co-pilot
and an Elastic/LongChain initiative \cite{ElasticAIrepo}
\label{foot:elastic}} have been identified as promising approaches.

An alert is just one observation and needs to be put into a larger context to identify an \emph{incident} and provide necessary \emph{situational awareness} as a result of the analysis \cite{franke2022cyber}. 
Such contextualisation includes enriching alerts with relevant knowledge from previous incidents, common systems behaviour, infrastructure details, threats, assets, etc. 
The same attack -- or the same phase of a larger attack -- is likely to trigger many different alerts. Different ML techniques, particularly clustering,  have been studied to fuse or aggregate related alerts \cite{tobias_masteroppgave,Kotenko22}. In addition to understanding an incident and achieving situational awareness, contextualisation will also help a security analyst understand individual alerts.
Similar to challenge \ref{challenge:extractk}, contextualization of alerts will necessarily involve extracting symbolic representation from a vast amount of available (and typically unstructured) information.
A cyber attack conducted by an advanced adversary will, in most cases, manifest itself over several phases, creating a need to discover the relationships (between the alerts) across the different phases of an attack.
A common reference model to relate such phases is the \emph{cyber kill chain}, originally developed by Lockheed Martin and later refined into the \emph{unified cyber kill chain} \cite{pols2017unified}. Other formalisms that enable modelling different phases of attacks include 
\emph{MITRE Attack Flow} \cite{ATTACKFLOW} and the \emph{Meta Attack Language} (MAL) \cite{johnson2018meta}. Different approaches have been studied to relate the different phases, including symbolic approaches \cite{ou2005mulval}, AI planning \cite{amos2017efficient,miller2018automated}, knowledge graphs \cite{cs7202,Chetwin24},  
state machines \cite{wilkens2021multi}, clustering \cite{Haas18} and
statistics \cite{haque2023adversarial}. However, this research topic is considerably less mature compared with ML models for detection in the monitor phase.
We summarise the challenges of combining, understanding and explaining observations in the following challenge:
\begin{challenge}
\label{challenge:alertctxt}\label{challenge:multistep}
Combine observation with knowledge to analyse, develop and communicate situational awareness.
\end{challenge}
\noindent Developing cyber situational awareness requires connecting a plethora of different sources, such as alerts and details about infrastructure and threats. 
There have been proposals to use knowledge graphs to combine these different sources to support analysis \cite{sikos2023cybersecurity,liu2022review}, including explanation \cite{bushra_a_alahmadi_99_2022}.

 When an incident is understood and sufficient situational awareness is achieved, a suitable amount of resources have to be allocated to handle the incident. There may be multiple incidents, requiring some prioritisation between them. This involves understanding the risks and potential impacts of the incident, including the risks and impacts of any mitigating actions that may be taken in subsequent MAPE-K phases:
\begin{challenge}
\label{challenge:riskimpact}
Understanding the risk, impact, importance and priority of incidents.
\end{challenge}

\paragraph{Plan \& Execute}

The last two phases of MAPE-K, plan and execute, focus on responding to detected incidents.
This involves finding suitable responses in the plan phase, and prepare and execute the response(s) in the execute phase. From an AI perspective, research in these phases is not as mature as in the monitor and analyse phases. We will here only focus on the plan phase, which we currently consider to have the more interesting AI-related challenges.
To plan a suitable response, two promising AI techniques are 
\emph{AI planning} (e.g.\ \cite{ghosh2012planner}) and \emph{reinforcement learning} (RL -- e.g.\ \cite{hu2020adaptive,nyberg2022cyber}). Each of them have their pros and cons: AI planning requires considerable knowledge and formulation of the underlying environment, whilst reinforcement learning requires a considerable amount of interactions/simulations (often in the millions). In certain cases, a quick response time is necessary, which means this level of interaction would be too time-consuming. When generating response actions, their risk and impact must be taken into account (including the risk and impact of doing nothing), which is an unsolved problem when using AI. Moreover, when proposing a response action, an AI-generated solution must be able to explain both what the response action will do and why it is suitable for the given problem:
\begin{challenge}
\label{challenge:responserisk}
Generate and recommend suitable response actions in a timely manner that take into account both risk and impact and are understandable for a security analyst.
\end{challenge}
\noindent To support such generation, there are several frameworks and formal ontologies that can be used, such as
MITRE \emph{D3FEND}~\cite{kaloroumakis2021toward}, \emph{RE\&CT} \cite{REACT} and \emph{CACAO playbooks} \cite{mavroeidis2022cybersecurity}.






\paragraph{Shared knowledge}
The `K' in MAPE-K stands for \emph{knowledge} shared across the phases, and we have, for instance, seen knowledge about threats and the infrastructure being protected used across different phases.
Moreover, this knowledge takes different forms and representations (structured and unstructured) and is analysed using different techniques (symbolic and sub-symbolic).
In addition to consuming knowledge, it is also important to share knowledge with key stakeholders, both technical and non-technical \cite{enisathreat22}. This may be a report about an incident for internal use (e.g.\ to board members) or sharing of discovered threat intelligence with a wider community:
\begin{challenge}\label{challenge:report}
Generating suitable incident and CTI reports for the target audience.
\end{challenge}


To summarise, we have shown the need to learn and reason across MAPE-K and that both symbolic and connectionist AI are being used across the phases. We have identified several challenges, and next, we make the case for NeSy to address them.

\section{The case for neurosymbolic AI}\label{sec:why}

Kahneman's \cite{daniel2017thinking} distinction between  (fast) instinctive and unconscious `\emph{system 1}' processes from  (slow) more reasoned `\emph{system 2}' processes, has often been used to illustrate the NeSy integration of neural networks (system 1) and logical reasoning (system 2). Building on this analogy, system 1 can, in a SOC, be seen as the ML-based AI used to identify potentially malicious behaviour in the monitor phase. Here, a large amount of noise needs to be filtered out from the large amount of events (thus a need for speed and scalability). System 2 is the reasoning conducted in the analysis phase, which requires deeper insight with the need for scalability less significant. 

This dichotomy of requirements entails that neither end-to-end pure statistical nor pure logical approaches will be sufficient, and a NeSy combination seems ideal.
Three commonly used reasons for pursuing NeSy are  to design systems that are 
\emph{human auditable and augmentable}, can \emph{learn with less} and provide \emph{out-of-distribution generalisation} \cite{grayWelcome2023}. We have seen examples of each of these in the challenges described in \S \ref{sec:problem}:
the use of knowledge to both contextualise, analyse and explain alerts, and to generate and explain and response actions; to learn from (relatively few) incidents; and to handle concept drifts and noise in order to achieve high accuracy of ML models under real-world conditions.
From the challenges in \S \ref{sec:problem}, we will here outline a set of NeSy use cases we believe are promising and identify some promising NeSy tools and techniques for each of them. This work is not complete and should be seen as a start (see \S \ref{sec:concl}). Moreover, this section is speculative by nature, but we provide some evidence in terms of existing work and experiments conducted in \S \ref{sec:experiments}.

\paragraph{Monitor}


The ability to integrate relevant knowledge into ML-based detection models (challenge \ref{challenge:extractk})
falls directly under the NeSy paradigm, and could both improve performance under real-world conditions (challenge \ref{challenge:performance}) and help to reduce the number of false alerts (challenge \ref{challenge:volume}): 
\begin{usecase}
Use (symbolic) knowledge of threats and assets to guide or constrain ML-based detection engines.
\label{usecase:guideTraining}
\end{usecase}
\noindent A similar case for such a NeSy use case is made in \cite{piplai2023knowledgeIEEE}.
\emph{Logical Neural Networks (LNN)} \cite{riegel2020logical} are  
designed to simultaneously  provide key properties of both neural nets (learning) and symbolic logic (knowledge and 
reasoning), enabling both logic inference and injecting desired knowledge (e.g.\ about threats and infrastructure) into 
the neural architecture.
In \emph{Logic Tensor Networks (LTN)} \cite{badreddine2022logic}, a membership function for
 concepts are learned based on both labelled examples and abstract (logical) rules.
  LTN introduces a fully differentiable logical language, called \emph{real
 logic}, where elements of first-order logic can be used to encode the necessary knowledge.
LTN has been studied to detect suspicious behaviour \cite{onchis2022advantages} and is the topic of one of our experiments in \S \ref{sec:experiments}.
 
 

In challenge \ref{challenge:smalldata}, we highlighted the need to learn from (relatively small) datasets, which is one of the key features of NeSy \cite{grayWelcome2023}:
\begin{usecase}
Learn detection models from a limited number of (labelled) incidents
\end{usecase}
\noindent Additional embedded knowledge in an LNN or LTN may help to reduce training time. \emph{NS-CL} \cite{mao2019neuro}, which builds models to learn visual perception including semantic interpretation of the images, has shown it can be trained on a fraction of the data required by comparable methods -- albeit in a different domain with different data sources. NeSy-based inductive logic programming variants, such as $\partial$\emph{ILP}  \cite{evans2018learning}, would also be able to learn from small datasets. The learned logic program will also be inherently explainable (see challenge \ref{challenge:alertctxt}).



Threat hunting involves generating suitable hypotheses, applying and validating them, then
update and iterate (challenge \ref{challenge:hunting}). 
Work has started investigating LLMs for this challenge \cite{Perrina23}. It has been argued
for symbolism in LLMs  \cite{hammond2023large}, and based on that we define 
an LLM-based NeSy threat hunting use case:
\begin{usecase}\label{usecase:hunting}
LLM-driven threat hunting using symbolic knowledge and reasoning capabilities.
\end{usecase}
\noindent
LLMs have been used for hypothesis generation in other domains \cite{qiu2023phenomenal}, which can be further investigated for threat hunting.
Hypothesis generation is typically driven by CTI, which can be captured in a knowledge graph. The integration of LLMs and knowledge graphs is an active research field \cite{Pan24}. In addition, symbolic or computational methods could be used for other steps in the hunting process, including: planning how to answer the hypothesis; reasoning about available data sources to execute this plan; 
ensuring correct translation to required query language\footnote{Events are stored in a SIEM system, which will have a query language.} to validate the hypothesis using the observations; and finally, reason about the results from the execution and provide input for any refinement of the hypothesis for a new hunting iteration.


\paragraph{Analyse}

A prominent characteristic of NeSy is its capacity to combine learning and reasoning. Such a combination is desirable in a SOC, and our next use case, which cuts across the monitor and analyse phases,  addresses several of the challenges from \S \ref{sec:problem}:  
\begin{usecase}
\label{uc:learnreason}
Incorporate learning of detection models with the ability to reason about their outcomes to understand and explain their nature and impact.
\end{usecase}
\noindent In \cite{piplai2023knowledgeIEEE}, the case for such integration of detection and analysis using NeSy is also made.
One way to achieve this is to simultaneously train a neural network (for detection) with related symbolic rules that can be used for contextualization, analysis and explanation (challenge \ref{challenge:alertctxt}). Two NeSy techniques that can accomplish this are \emph{Deep Symbolic Learning (DSL)} \cite{daniele2023deep}
and \emph{Neuro-Symbolic Inductive Learner (NSIL)} \cite{cunnington2023neuro}.
\emph{dPASP} \cite{geh2023dpasp} and \emph{NeurASP} \cite{yang2023neurasp} are techniques based on Answer Set Programming (ASP) \cite{brewka2011answer}, which seems promising for this use case. \emph{dPASP} is based on furnishing ASP with neural predicates as interface to both deep learning components and probabilistic features in order to afford differentiable neurosymbolic reasoning. dPASP is suitable for detecting under incomplete information, abductive reasoning, analysis of competing hypotheses (ACH) \cite{heuer1999analysis}, and what-if reasoning.\footnote{dPASP can leverage existing LLM-ASP integrations \cite{rajasekharanReliableNaturalLanguage2023} to facilitate use case \ref{usecase:hunting}.}
\emph{NeurASP} improves the results from neural classifiers by applying knowledge-driven symbolic reasoning to them. This is achieved by treating the output from the neural classifier as a probability distribution over the atomic facts, which are then treated in ASP. The ASP rules can also be used to improve the training of the neural networks (use case \ref{usecase:guideTraining}).
\emph{DeepProbLog}~\cite{manhaeve2018deepproblog} and \emph{DeepStochLog}~\cite{winters2022deepstochlog}  incorporate reasoning, probability and deep learning, by extending probabilistic logic programs with neural predicates created from a neural classifier, and may thus provide the necessary combination of learning and reasoning for this use case.

Breaking use case \ref{uc:learnreason} into smaller sub-cases, the first being extracting symbolic alerts, in order to support alert contextualization, analysis and explanation:
\begin{usecase}\label{usecase:symbolicalert}
Extracting alerts in a symbolic form.
\end{usecase}
\noindent 
In \cite{himmelhuber2022detection}, symbolic alerts are extracted from a 
\emph{graph neural network} (GNN) based detection engine. A combination of \emph{GNNExplainer} \cite{ying2019gnnexplainer} and DL-Learner \cite{lehmann2009dl} is used to extract the symbolic alerts.
The symbolic rules learnt by both DSL and NSIL may also provide such symbolic alert representation, and the use of e.g. $\partial$\emph{ILP} for detection will learn symbolic alerts by design. 
\emph{Embed2Sym} \cite{aspis2022embed2sym} extracts latent concepts from a neural network architecture, and assigns symbolic meanings to these concepts. These symbolic meanings can then be used to encode symbolic alerts.

A  SOC typically receives a large volume of threat intelligence, which is too large to thoroughly analyse manually. Such intelligence is used to contextualise alerts, and it is thus desirable to enrich the SOCs knowledge bases with relevant intelligence reports:
\begin{usecase}
\label{usecase:statsenrich}
Use statistical AI to enrich or extract symbolic knowledge. 
\end{usecase}
\noindent This use case addresses challenge \ref{challenge:alertctxt}. In \S \ref{sec:problem}, we discussed several approaches to extract knowledge in a suitable symbolic form from reports \cite{marchiori2023stixnet,li2022attackg,liu2023constructing}. 
 \emph{STAR} \cite{rajasekharanReliableNaturalLanguage2023} is a possible NeSy technique that can be used. It combines LLMs with ASP, where ASP can be employed to reason over the extracted knowledge, in addition to just extracting it.

This ability to reason is crucial as the intelligence report may be incorrect or superseded for different reasons, including underlying (aleatoric) uncertainty, deterioration over time, or from sources one does not fully trust. It may also simply not be relevant for our purposes, or more importantly, intelligence reports may conflict with our existing knowledge or our observations.
It would therefore be desirable to quantify and reason about knowledge, 
including the level of trust, from both own observations and existing knowledge:
\begin{usecase}
Reason about and quantify knowledge. 
\end{usecase}
\noindent
This use case is aimed at addressing challenge \ref{challenge:riskimpact}. It may play a role in implementing a technique known as \emph{risk-based alerting} \cite{SplunkRBA}, which involves using data analysis to determine the potential severity and impact of alerts and incidents. 
\emph{Probabilistic attack graphs} \cite{gylling2021mapping} has been used to add probabilities to CTI.  One potential NeSy approach for this use case is  \emph{Recurrent Reasoning Networks} (RRNs)
\cite{hohenecker2020ontology}. RRNs could be used to train a ML model from observations to reason about our existing knowledge graph, e.g.\ to quantify or identify inconsistencies. Another NeSy example is \emph{Neural Probabilistic Soft Logic} (NeuPSL) \cite{ijcai2023p461}, where the output from the trained neural networks is in (symbolic) \emph{Probabilistic Soft Logic} (PSL) \cite{bach2017hinge}, which can be treated by probabilistic graphical models.
NeurASP, dPASP, DeepProbLog and DeepStochLog may also be applicable here.


As discussed in \S \ref{sec:problem}, a cyber attack conducted by an advanced adversary will consist of multiple phases and the ability to relate these phases is essential when developing cyber situational awareness (challenge \ref{challenge:multistep}):
\begin{usecase}\label{usecase:phases}
Relate the different phases of cyber incidents. 
\end{usecase}
\noindent One concrete NeSy use case would be to merge the statistics- and semantics-driven approaches outlined in \cite{Applebaum19}. 
Further, \emph{PyReason} \cite{aditya2023pyreason} enables temporal reasoning over graphical structures, such as knowledge graphs. 
This can be used to exploit the temporal aspect of relating the different phases. 
The second experiment in \S \ref{sec:experiments} addresses this use case by
exploring temporal reasoning using a combination of LLMs, temporal logic, ASP and plan recognition. 
The ontological reasoning supported by RRNs also seems promising for this type of problem. 




\paragraph{Plan \& execute}

\emph{Neurosymbolic reinforcement learning} (NeuroRL) \cite{Acharya23} combines the respective advantages of reinforcement learning and AI planning. NeuroRL can learn with fewer interactions compared with traditional RL by using inherent knowledge, thus making it more applicable than both RL and AI planning when (near) real-time response is required and a complete model of the environment is infeasible. Moreover, it has the promise of more explainable response actions, whilst a reasoning engine could, in principle, help to take into account both risk and impact\footnote{We note that there are additional challenges when generating risk and impact-aware responses, such as both deriving the requirements in the first place and representing them in a suitable way.}. Thus, this seems like promising approach for challenge \ref{challenge:responserisk}:
\begin{usecase}
Generating impact and risk aware explainable response actions in a timely fashion using neurosymbolic reinforcement learning,
\end{usecase}
\noindent Neurosymbolic reinforcement learning has been used in offensive cyber security settings for penetration testing\footnote{Penetration tests are simulated attacks against the infrastructure and assets being protected, for instance, to identify vulnerabilities.} \cite{dingaccelerating24}. Whilst there are some commonalities with our challenges, defending has their own peculiarities.
For example, speed, risk, impact and explainability are more prominent when defending against attacks.

\paragraph{Shared knowledge}


Our final use case directly addresses challenge \ref{challenge:report}. 
LLMs are extensively studied for generating reports and this is also the
case for cyber defence \cite{motlagh2024large}. The generation process is likely to use symbolism (e.g.\ knowledge graphs \cite{Pan24}). The reports need to be correct, which is one area symbolic AI can help \cite{hammond2023large}. We, therefore, rephrase challenge \ref{challenge:report} as a NeSy use case:
\begin{usecase}
Generation of incident reports and CTI reports tailored for a given audience and/or formal requirements, using (symbolic) knowledge and LLMs.
\end{usecase}

\section{Proof-of-concept experiments}\label{sec:experiments}

To showcase the feasibility of using NeSy for our use cases, we have conducted two initial experiments: 1) using LTN to show how knowledge in symbolic form can be used to improve an ML-based detection engine (use case \ref{usecase:guideTraining}); and 2) using LLMs and ASP to elicit and reason with adversary attack patterns and observed alerts for situational awareness (use case \ref{usecase:statsenrich}).
Both experiments are deliberately simplified and are not conducted in realistic conditions. They do, however, demonstrate the potential of NeSy in a SOC setting. Additional technical details can be found in  appendix \ref{app:details} and all source code can be found on GitHub: \texttt{\url{https://github.com/FFI-no/Paper-NeSy24}}.

\paragraph{Experiment 1: LTN for knowledge-aware intrusion detection}\label{sec:experiments:ltn}


The first experiment addresses use case \ref{usecase:guideTraining} and is in the monitor phase of MAPE-K. Here, the goal is to detect malicious traffic by training a classifier that can separate benign traffic from two different classes of malicious traffic: brute force attacks and cross-site scripting (XSS) attacks. The classifier will produce an alert if malicious traffic is seen. The input data are \emph{NetFlow} entries \cite{netflow2024cisco}, which provide aggregated information on traffic between two distinct ports on distinct IP addresses for a given protocol. The data is a subset of the CICIDS2017 dataset \cite{sharafaldin2018toward}, which is a labelled dataset containing simulated attacks on a network, with additional details in appendix \ref{app:details}.

The experiment consists of two parts: In the first part, a simple 3-layered fully connected neural network is trained and used as a baseline. In the second part, this neural network is extended using an LTN \cite{badreddine2022logic}, where additional domain knowledge is encoded. In both cases, $70\%$ of the data is used for training and $30\%$ for testing. The experiment is inspired by \cite{onchis2022advantages}, where LTN is used in a similar, but more limited, way. 

The LTN consists of one predicate for class membership, $P(x,class)$,
and is configured as a neural network with the same structure as the baseline neural network. 
We define the following axioms (expressed in Real Logic \cite{badreddine2022logic}):
$$ \forall x \in {B} : P(x, Benign) \quad
\forall x \in {BF} : P(x, Brute\_force) \quad
 \forall x \in {X} : P(x, XSS)$$
These axioms describe how all flows in the training set labelled as a given class are a member of the class. This encodes the information of the baseline neural network with no additional knowledge. 
From the dataset network topology, we define \textit{NWS} to be the set of all NetFlows that do not communicate with a web server. We know that if a NetFlow is in this \textit{NWS}-set, then it cannot be a web attack (that we are interested in). We add this domain knowledge as a fourth axiom used by the LTN:
$$ \forall x \in {NWS} : \neg (P(x, Brute\_force) \vee P(x, XSS))$$
Training consists of updating the neural network $P$ to maximize the accumulated truth value of the axioms~\cite{badreddine2022logic}. 
The following table shows the results on the test data:

\begin{center}
\begin{tabular}{|l|rrr|rrr|}
\hline
 & \multicolumn{3}{l|}{\textbf{Baseline Neural Network}} & \multicolumn{3}{l|}{\textbf{Logic Tensor Network}} \\ \hline
\textbf{Labels}      & \multicolumn{1}{l|}{\textbf{Precision}} & \multicolumn{1}{l|}{\textbf{Recall}} & \textbf{F1}    & \multicolumn{1}{l|}{\textbf{Precision}} & \multicolumn{1}{l|}{\textbf{Recall}} & \textbf{F1}    \\ \hline
Benign      & \multicolumn{1}{l|}{0.999}     & \multicolumn{1}{l|}{0.916}  & 0.956 & \multicolumn{1}{l|}{0.998}     & \multicolumn{1}{l|}{0.963}  & 0.981 \\ \hline
Brute Force & \multicolumn{1}{l|}{0.090}     & \multicolumn{1}{l|}{0.686}  & 0.159 & \multicolumn{1}{l|}{0.155}     & \multicolumn{1}{l|}{0.622}  & 0.248 \\ \hline
XSS         & \multicolumn{1}{l|}{0.088}     & \multicolumn{1}{l|}{0.628}  & 0.155 & \multicolumn{1}{l|}{0.213}     & \multicolumn{1}{l|}{0.648}  & 0.321 \\ \hline
\end{tabular}
\end{center}

This was a very simple extension, yet precision was almost doubled, illustrating the potential for using NeSy to embed additional knowledge in ML models. Recall was largely the same.
This was expected since the additional axiom was related to removing false alerts and not to find missed attacks. Further knowledge could be additional details of our infrastructure or assets (including known vulnerabilities) as well as information about the attacker (e.g., from threat intelligence). Furthermore, state-of-the-art detectors are not as simple as the neural network used here and will typically combine related NetFlows before detecting (see, e.g., \cite{computers10060079}). Still, this simplified version shows promise for the use of NeSy to enrich ML-based models with (symbolic) knowledge.

\paragraph{Experiment 2: LLMs and ASP for situational awareness}

The second experiment illustrates the use of NeSy to relate different phases of an attack
(use case \ref{usecase:phases}). Here, alerts that are sequenced by time, are mapped to adversary attack patterns, gleaned from textual CTI reports into symbolic form using statistical methods (use case \ref{usecase:statsenrich}). The experiment is inspired by existing work such as: neurosymbolic plan recognition \cite{amado2023robust}, attack plan recognition \cite{amos2017efficient}, and the use of LLMs to extract both LTL \cite{fuggitti2023nl2ltl} and CTI (in the form of MITRE ATT\&CK \emph{tactics} or \emph{techniques})\footnote{A MITRE ATT\&CK \emph{tactic} describes why
an adversary performs an action, while a MITRE ATT\&CK \emph{technique} describes how the action is performed \cite{ATTACK}.} \cite{haque2023adversarial,orbinato2022automatic,you2022tim}. 
An LLM (GPT4) is first used to elicit formal representations of attack patterns described in CTI reports, affording us a rapid way of converting CTI to symbolic knowledge. Here, we use the \texttt{NL2LTL} Python library \cite{fuggitti2023nl2ltl} to extract representations of attack patterns in $LTL_{f}$ \cite{de2013linear}, a temporal logic for finite traces. 
 The following visualises a conceptual adversary attack pattern, sequencing MITRE ATT\&CK techniques: 
 
\begin{center}
\begin{tikzcd}[cramped, sep=tiny]   
&\{{t1556}\} & \ldots & \{{t1059}\} & \ldots & \{{t1548}\} & \ldots & \{{t1059}\}\\
\arrow[r] &\bullet \arrow[r,""] & \ldots \arrow[r] & \bullet \arrow[r]  & \ldots \arrow[r] & \bullet \arrow[r] & \ldots \arrow[r] &\bullet\\
&t_{i} & \ldots & t_{j} & \ldots & t_k &\ldots&t_{l}\\
\end{tikzcd} \linebreak
$LTL_f$ : $\square(\mathbf{I}\rightarrow\circ\lozenge ({t1556} \wedge \circ\lozenge ({t1059} \wedge \circ \lozenge ({t1548} \wedge \circ \lozenge {t1059}))))$
\end{center}

\noindent Each `\emph{txxx}', where $x$ is a number, is a unique technique from the MITRE ATT\&CK framework \cite{ATTACK}, $\mathbf{I}$ is the initial state, and $\square$, $\circ$ and $\lozenge$ are the `always', `next' and `eventually' operators in LTL. Next, {\texttt{telingo}\cite{cabalar2018temporal}, an ASP solver for temporal programs,  is used to postdict possible attacks. It is given: (i) $LTL_{f}$ representations of known attack patterns; (ii) sequences of observed alerts; and (iii) knowledge linking alerts to techniques. 
The attack patterns outlined in (i) are acquired by the elicitation step described above, and the sequences of observed alerts outlined in (ii) are assumed to come from a SIEM system. That is, alerts produced are in a structured form amenable to be represented as Prolog/ASP terms. We assume that this conversion of alerts to symbolic form (use case \ref{usecase:symbolicalert}) exists (see e.g.  \cite{himmelhuber2022detection}). Furthermore, they are temporally ordered inducing a sequence of alerts (where $a_{x}$ is an alert in symbolic form): 
\begin{center}
\begin{tikzcd}[cramped, sep=tiny]  
\{a_{addGrpMem}\} & \{a_{benign}\} & \{a_{execIam}\} & \ldots & \{a_{latMvmSaml}\} & \{a_{benign}\} & \{a_{execWinPsh}\}\\
\bullet \arrow[r,""] & \bullet \arrow[r] & \bullet \arrow[r]  & \ldots \arrow[r] & \bullet \arrow[r] & \bullet \arrow[r] &\bullet\\
t_1 & t_2 & t_3 & \ldots & t_{n-2} & t_{n-1} & t_{n}\\
\end{tikzcd}
\end{center}
Finally, 
we assume that all the alerts produced can be associated with MITRE ATT\&CK techniques, which is common for many signature-based alerts. Note, however, that it is a many-to-many relationship: an alert can be an indicator for several techniques, and a technique can have several alert indicators. This knowledge\footnote{Extracted from alert rules found at \texttt{\url{https://github.com/SigmaHQ/sigma}}.} can be represented in ASP with choice rules: 
\begin{center}
\[
\begin{array}{l r}
1 \: \{t1556;t1548\} \: 1 \leftarrow a_{addGrpMem} & \\
1 \: \{t1059\} \leftarrow a_{execIam} & \\
1 \: \{t1548\} \leftarrow a_{latMvmSaml} & \\
1 \: \{t1059\} \leftarrow a_{execWinPsh} & \\
\end{array}
\]
\end{center}
We encode the problem in a \texttt{telingo} program where the adversary attack plan from (i) is formulated as a model constraint, the trace in (ii) is encoded as a temporal fact, and the rules in (iii) are encoded as dynamic rules. A stable model produced by the \texttt{telingo} solver tells us that it is plausible that the trace is an instance of the attack plan, while the absence of a model will rule it out, thus achieving the goal set out in the experiment. The \texttt{telingo} program and LTL extraction are detailed in appendix \ref{app:details}. 
\section{Conclusion}
\label{sec:concl}

Our main goal with this paper has been to showcase the possibilities for NeSy in cyber security, focusing on problems within SOCs, which we hope will help stimulate a concerted effort in studying NeSy in this domain. The use of NeSy in cyber security is in its infancy, with some work 
having appeared over the last few years, including using for detection \cite{onchis2022advantages}, generating symbolic alerts \cite{himmelhuber2022detection} and extracting semantic knowledge from reports \cite{marchiori2023stixnet}. 

We have demonstrated that a considerable amount of symbolic and statistics-based AI is studied in SOC settings, and using it in real-world settings presents several challenges. We believe NeSy can address many of these challenges. Others have made some of the same points \cite{jalaipan23,piplai2023knowledgeIEEE}, but not to the extent as we do here.
We have contributed by defining a set of NeSy use cases and identifying promising NeSy approaches that serve as a starting point—two of them demonstrated in our proof-of-concept experiments. This work is just a start, and we both hope and expect that many new use cases and promising NeSy approaches that we have not covered here will appear in the not-too-distant future. 
The use cases and experiments provide a starting point for such future work. A challenge with AI in the cyber security domain is available datasets. Due to issues such as privacy, confidentiality and lack of ground truth, researchers tend to use synthetic data, which has their limitations \cite{kenyon2020public,apruzzese2023sok}. Furthermore, such datasets tend to focus only on detection (monitor phase), containing only events and lack the additional (symbolic) knowledge which is important in SOCs and for our use cases. This is also one of the reasons that our experiments lack realistic conditions. An important first step will be to develop synthetic datasets containing both events for detection and necessary knowledge in order to address the use cases. This can either be achieved by extending existing ``detection datasets" \cite{kilincer2021machine} with the necessary knowledge or by developing new  ``NeSy datasets" from scratch.

\paragraph{Acknowledgements} This work was partly funded by the European Union as part of the European Defence Fund (EDF) project AInception (GA No. 101103385). Views and opinions expressed are however those of the authors only and do not necessarily reflect those of the European Union (EU). The EU cannot be held responsible for them.

\bibliographystyle{plain}

\bibliography{refs}

\appendix

\section{Further details of experiments}\label{app:details}

\subsection{Experiment 1}

The recorded NetFlows are an aggregation of metadata of the IP traffic in the network. All traffic where a specific subset of features are the same is believed to be regarding the same activity is recorded as the one flow~\cite{netflow2024cisco}.

The dataset is a subset of the CICIDS2017 dataset~\cite{sharafaldin2018toward}, specifically looking at the "Tursday Morning" part. For simplicity we have chosen a selection of flow features and attack classes. The datasets contain labelled flows categorized into the categories: Web Attack - Benign, Web Attack – Brute Force, and Web Attack - XSS. The classes are greatly imbalanced, with 168 000 flows in the benign class and 2 159 flows in the remaining classes. Such imbalance is common 
for this type of cyber security dataset where most traffic (by far) will be benign. The dataset is partitioned into a $30\%$/$70\%$-split between the training set and the test set. Each flow is encoded as a tensor.

The NetFlow features extracted in this experiment are:
\begin{multicols}{2}
\begin{itemize}
    \item Source Port
    \item Destination Port
    \item Protocol
    \item Flow Duration
    \item Total Length of Fwd Packets
    \item Total Length of Bwd Packets
    \item Fwd Header Length
    \item Bwd Header Length
\end{itemize}
    
\begin{itemize}
    \item Fwd PSH Flags
    \item FIN Flag Count
    \item Bwd Packet Length Min
    \item Init Win bytes forward
    \item Init Win bytes backward
    \item Subflow Fwd Bytes
    \item Total Length of Fwd Packets
\end{itemize}
\end{multicols}

The baseline neural network is encoded as a fully connected multi-layer perceptron with dimensions (16,16,8). This represents the predicate for class membership $P(x,class)$. The samples are assigned to the class with the highest confidence.

The weights of the network are updated using the Adam optimization algorithm~\cite{kingma2014adam} with the aggregated truth of axioms used to define the loss. The network is trained over 200 epochs.

\subsection{Experiment 2}

\subsubsection{\texttt{telingo} program}
The following code listing shows the \texttt{telingo} program from the first experiment:

\begin{lstlisting}%[language=Prolog]

#program initial.
1 { plan(plan1;plan2) } 1 .

%Trace of alerts
&tel{&true
    ;> o(alert(addGrpMem))
    ;> o(alert(benign)) 
    ;> o(alert(execIam))
    ;> o(alert(latMvmSaml))
    ;> o(alert(benign)) 
    ;> o(alert(execWinPsh))}.

% Adversary attack plans
:- plan(plan1), not &tel{
    >? (h(techn(t1556)) 
    & > (>? h(techn(t1059)) 
    & > (>? h(techn(t1548))
    & > (>? h(techn(t1059))))))}.

:- plan(plan2), not &tel{
    >? (h(techn(t1556)) 
    & > (>? (h(techn(t1059)) | h(techn(t1548)))))}.

#program dynamic.

% abduce technique based on alert
1 {hyp(techn(t1556)); hyp(techn(t1548))} 1 :- o(alert(addGrpMem)) .
1 {hyp(techn(t1059))} 1 :- o(alert(execIam)) .
1 {hyp(techn(t1548))} 1 :- o(alert(latMvmSaml)) .
1 {hyp(techn(t1059))} 1 :- o(alert(execWinPsh)) .
 
%hypothesized -> happened
h(X) :- hyp(X).
\end{lstlisting}

\subsubsection{CTI transformed to LTL}
This subsection explains how CTI reports of previous attacks are transformed into LTL temporal representations of the attack patterns. The CTI reports are in natural language and we utilize the \texttt{NL2LTL}~\cite{fuggitti2023nl2ltl} Python package for the translation. We define a custom pattern template $ExistenceEventuallyOther$ to express the LTL property $ \lozenge a \wedge \circ \lozenge b$. Additionally, we create a custom prompt tailored for our domain. The prompt contains the allowed pattern ($ExistenceEventuallyOther$), allowed symbols (MITRE ATT\&CK technique IDs), and multiple examples.
The LTL representation is a chain of MITRE ATT\&CK techniques. However, CTI descriptions of an attack might not reference any specific techniques. In those cases, we let the LLM deduce which MITRE ATT\&CK technique is referenced from the general description in natural language. This is a minimal implementation, and extraction of MITRE ATT\&CK \emph{tactics} from CTI reports have been investigated before~\cite{orbinato2022automatic,you2022tim}.


\begin{lstlisting}[caption={Prompt}, captionpos=b]
Translate natural language sentences into patterns:
ALLOWED_PATTERNS: ExistenceEventuallyOther
ALLOWED_SYMBOLS: T1548 (Abuse Elevation Control Mechanism),
T1530 (Data From Cloud Storage), [...]

NL: The adversary logs into the Kubernetes console. 
This leads to: The adversary can view plaintext AWS keys 
in the Kubernetes console.
PATTERN: ExistenceEventuallyOther
SYMBOLS: T1133, T1552
[...]
\end{lstlisting}

A previous attack is described in natural language, where it is assumed that the first sentence describes an action that happened before the second sentence. 
The sentences in Listing \ref{lst:attack} correspond to the MITRE ATT\&CK tactics T1566, T1548 and T1048, respectively. This description is based on the procedure examples of techniques in MITRE ATT\&CK~\cite{ATTACK}.

\begin{lstlisting}[caption={attack description}, captionpos=b, label=lst:attack]
Attackers leveraged spearphishing emails with malicious links
to gain access to the system. Attackers modifies the tty_tickets
line in the sudoers file to gain root access. Exfiltration over 
standard encrypted web protocols to disguise the exchanges as
normal network traffic.
\end{lstlisting}
We combine each sentence with the next sentence by adding \emph{``This leads to: "}. The sentences are given to \texttt{NL2LTL}. As a result, we get a sequence of LTL statements that can be combined into one statement. In this example, the three sentences are transformed into the following LTL formula:
$$\square(\mathbf{I}\rightarrow\circ\lozenge(t1566 \wedge \circ \lozenge (t1548 \wedge \circ \lozenge t1048)))$$

\end{document}